\PassOptionsToPackage{table}{xcolor}
\documentclass[conference,a4paper]{Latex-2025/APSIPA2025}
\usepackage{amsmath}
\usepackage{graphicx}
\usepackage{multirow}
\usepackage{threeparttable}
\usepackage{booktabs}
\usepackage{caption}
\definecolor{gray}{rgb}{0.4, 0.4, 0.4}
\definecolor{clsTshirt}{rgb}{0.200,0.133,0.533}   
\definecolor{clsTrouser}{rgb}{0.533,0.800,0.933}  
\definecolor{clsPullover}{rgb}{0.067,0.467,0.200} 
\definecolor{clsDress}{rgb}{0.867,0.800,0.467}    
\definecolor{clsCoat}{rgb}{0.800,0.400,0.467}     
\definecolor{clsSandal}{rgb}{0.667,0.267,0.600}    
\definecolor{clsShirt}{rgb}{0.267,0.667,0.600}    
\definecolor{clsSneaker}{rgb}{0.600,0.600,0.200}  
\definecolor{clsBag}{rgb}{0.533,0.133,0.333}      
\definecolor{clsAnkleboot}{rgb}{0.400,0.067,0.000}
\usepackage{graphicx}
\usepackage{subcaption}
\usepackage[margin=1in]{geometry}
\usepackage{caption}
\usepackage{algorithm}
\usepackage{algpseudocode}
\usepackage{algorithm}
\usepackage{multirow}
\usepackage{setspace}
\usepackage{amssymb}

\usepackage{geometry}
\usepackage{fancyhdr}

\renewcommand{\tablename}{Table}
\renewcommand{\thetable}{\arabic{table}}
\fancypagestyle{firststyle}{
  \fancyhf{}
}

\usepackage{paralist}

\begin{document}

\title{Reshaping the Forward-Forward Algorithm\\
with a Similarity-Based Objective}

\author{
\authorblockN{
James Gong\authorrefmark{1}\thanks{James Gong, Emma Wang, Raymond Luo, and Leon Ge contributed equally to this work.},
Raymond Luo\authorrefmark{1},
Emma Wang\authorrefmark{1},
Leon Ge\authorrefmark{1},
Bruce Li\authorrefmark{2},
Felix Marattukalam\authorrefmark{1},
and Waleed Abdulla\authorrefmark{1}
}

\authorblockA{\authorrefmark{1}University of Auckland, New Zealand \\
Emails: \{hgon777, rluo154, ewan353, sge524\}@aucklanduni.ac.nz,\\
Felix.Marattukalam@auckland.ac.nz, w.abdulla@auckland.ac.nz}

\authorblockA{\authorrefmark{2}Imperial College London, United Kingdom \\
Email: bruce.li25@imperial.ac.uk}

}
\maketitle
\thispagestyle{firststyle}
\pagestyle{fancy}

\begin{abstract}
  Backpropagation is the pivotal algorithm underpinning the success of artificial neural networks, yet it has critical limitations such as biologically implausible backward locking and global error propagation. To circumvent these constraints, the Forward-Forward algorithm was proposed as a more biologically plausible method that replaces the backward pass with an additional forward pass. Despite this advantage, the Forward-Forward algorithm significantly trails backpropagation in accuracy, and its optimal form exhibits low inference efficiency due to multiple forward passes required. In this work, the Forward-Forward algorithm is reshaped through its integration with similarity learning frameworks, eliminating the need for multiple forward passes during inference. This proposed algorithm is named Forward-forward Algorithm Unified with Similarity-based Tuplet loss (FAUST). Empirical evaluations on MNIST, Fashion-MNIST, and CIFAR-10 datasets indicate that FAUST substantially improves accuracy, narrowing the gap with backpropagation. On CIFAR-10, FAUST achieves \textbf{56.22\%} accuracy with a simple multi-layer perceptron architecture, approaching the backpropagation benchmark of \textbf{57.63\%} accuracy. 
\end{abstract}

\section{Introduction}
Artificial Neural Networks (ANNs), being simplified models of the brain, have achieved remarkable success across a wide range of tasks. A core algorithm enabling this success is backpropagation (BP) \cite{rumelhart1986learning}. BP uses the chain rule to compute the gradients of losses with respect to weights, facilitating gradient-descent-based optimization. Despite its central role, BP's reliance on global error signals, non-local updates, and backward locking are not biologically plausible \cite{Song_Lukasiewicz_Xu_Bogacz_2020}. Further, BP requires storing all intermediate activations during training, which can be restrictive for memory-constrained environments and parallelization capacity. These limitations have motivated the development of alternative biologically inspired training algorithms that rely on local learning signals and better relate to biological neural networks.\\\\
The Forward-Forward (FF) algorithm \cite{hinton2022forwardforwardalgorithmpreliminaryinvestigations} was proposed as an alternative learning algorithm that eliminates the need for BP. This approach avoids backward locking, and its local update nature aligns more closely with the principles of Hebbian learning \cite{Hebb_1949} and local plasticity \cite{STDP-3}. Its predictive accuracy, however, is lower than traditional BP. In this research, we reshape the core framework of FF by incorporating concepts of similarity learning \cite{schroff2015facenet}, leveraging an anchor-positive-negative structure to guide local learning dynamics. Our proposed algorithm is named \textit{Forward-forward Algorithm Unified with Similarity-based Tuplet loss} (\textit{FAUST}). We demonstrate that \textit{FAUST} achieves higher testing accuracy and improved inference efficiency compared to traditional FF on classification problems.

\section{Related Work}
\textbf{The Forward-Forward Algorithm.} FF, introduced as a biologically plausible alternative to BP, has inspired a variety of improvements and extensions. A key limitation of FF is its asymmetric treatment of positive and negative activations. To overcome this vulnerability, SymBa-FF \cite{lee2023symbasymmetricbackpropagationfreecontrastive} introduces a symmetric contrastive loss function that improves convergence speed; SFFA \cite{terresescudero2025contrastivesymmetricforwardforwardalgorithm}, instead, partitions each layer into dedicated positive and negative subsets and frames the algorithm within a continual learning paradigm. Subsequent efforts have extended FF beyond multi-layer perceptrons (MLPs) to convolutional neural networks (CNNs) \cite{dooms2023trifectasimpletechniquestraining, Papachristodoulou_2024, scodellaro2024trainingconvolutionalneuralnetworks}. Other approaches include collaborative FF (Collab FF)  \cite{Lorberbom_Gat_Adi_Schwing_Hazan_2024}, which enhances information flow across layers, and FF with
contrastive marginal loss (FFCM) \cite{10491157}, which augments input images to train an encoder with contrastive loss. Both strategies improve accuracy, and the latter is applied to vision transformers \cite{aghagolzadeh2025contrastiveforwardforwardtrainingalgorithm}. Our proposed approach lies within the same subcategory as \cite{10491157}, incorporating FF into a similarity learning framework.

\textbf{Similarity Learning.} Similarity learning involves training models through metrics such as the Euclidean distance and cosine similarity to produce embeddings, so that semantically similar inputs are closer together, and dissimilar ones are further apart. Siamese networks with contrastive loss \cite{contrastive} and triplet loss \cite{schroff2015facenet} have laid the foundation for this paradigm. Triplet-based training, however, often suffers from slow convergence. This limitation has led to improved sampling strategies (e.g., semi-hard mining \cite{hermans2017defensetripletlossperson}) and full-comparison losses that leverage multiple negatives per anchor (e.g., ${(N+1)}$-tuplet loss \cite{NIPS2016_6b180037}). Our method embeds FF into a similarity learning framework, using anchor-positive-negative relationships and ${(N+1)}$-tuplet loss to enhance the convergence and representation power of the algorithm. Additionally, naive implementations of the triplet and ${(N+1)}$-tuplet loss functions suffer from high computational complexity: training on all valid triplets has a runtime complexity of ${\mathcal{O}(M^3/C)}$, where $M$ is the size of the training set, and $C$ is the number of classes. To reduce the per-batch computation, \cite{snell2017prototypicalnetworksfewshotlearning, guerriero2018deepncm, 8954375, allen2019infinitemixtureprototypesfewshot} have proposed methods that approximate a representation for each class. This informative representation attempts to summarize each class and eliminate the need to compare all tuplets. We draw upon similar ideas in Section \ref{sec:representative-sec}.
\begin{figure*}[h]
    \centering
    \includegraphics[width=0.95\linewidth]{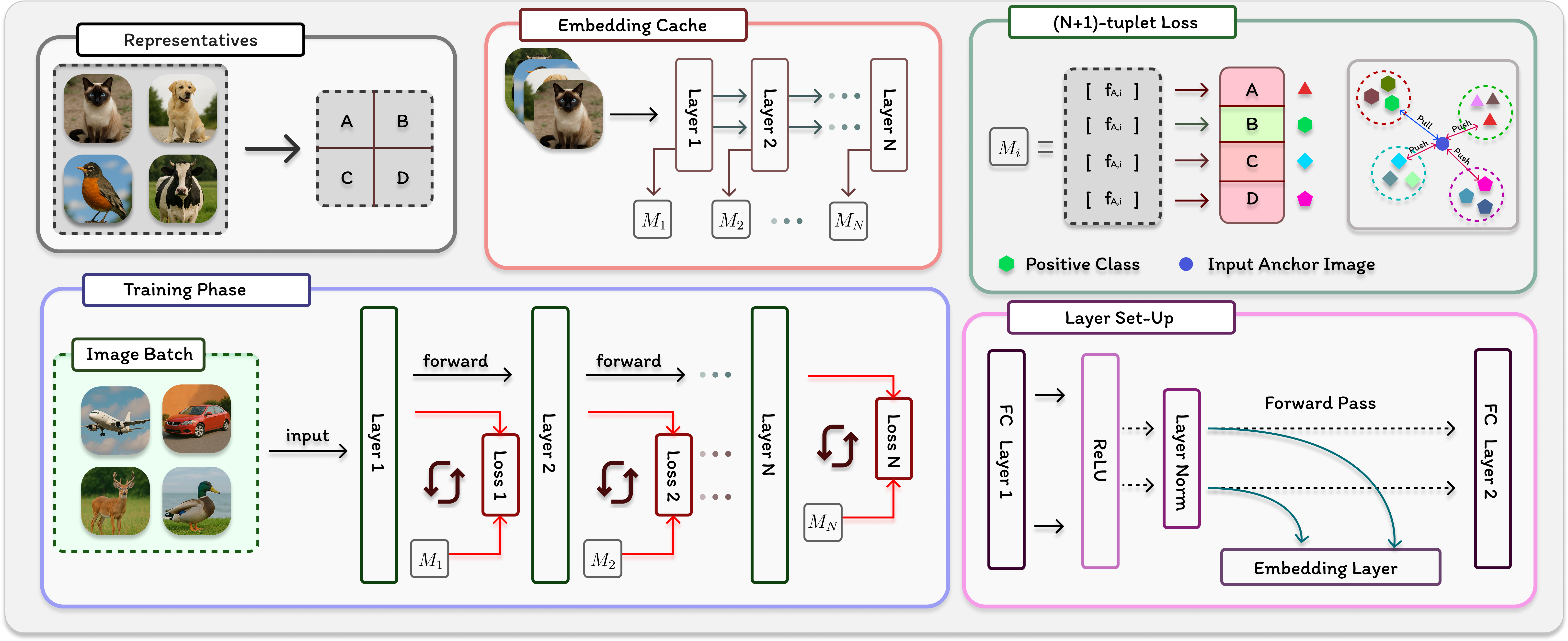}
    \caption{Overall algorithm architecture of \textit{FAUST}.}
    \label{fig:architecture}
\end{figure*}
\section{The Forward-Forward Algorithm}
FF replaces the back-and-forth of BP with two forward passes through each layer, one on `positive' data and one on `negative' data. In FF, the first \(C\) dimensions of the input vector are reserved for label encoding, where \(C\) is the number of classes. For a sample from class \(i\), the first \(C\) elements are set to a one-hot encoded vector with 1 at index \(i\) and 0 elsewhere. Positive data are constructed by assigning the correct one-hot encoded label, whereas negative data are constructed by assigning an incorrect one-hot encoded label. \\\\
During training, positive and negative data are passed through the current layer $i$, producing feature outputs $f^{\text{pos}}_i$ and $f^{\text{neg}}_i$. The `goodness' score, indicating the likelihood of the current encoding being correct, is computed for both outputs:
\begin{equation}
    G_i^{\text{pos}} = ||f_i^{\text{pos}}||^2,\;G_i^{\text{neg}} = ||f_i^{\text{neg}}||^2.
\end{equation}
The objective is to increase the goodness for positive inputs and reduce it for negative ones. The loss function is defined as
\begin{equation}
    \mathcal{L}^{\text{FF}}_i = \frac{1}{2}[\text{log}(1 + e^{\theta - G_i^{\text{pos}}}) + \text{log}(1 + e^{G_i^{\text{neg}} - \theta})],
\end{equation}
where $\theta$ is a tunable threshold shifting the margin between positive and negative activations. The weights in the current layer $i$ are updated via gradient descent, while the outputs $f_i^{\text{pos}}$ and $f_i^{\text{neg}}$ are normalized and forwarded as inputs to the next layer. This learning mechanism is applied throughout the network, allowing the model to train using only local error signals in a layer-wise manner.\\\\
In the inference phase, the model evaluates all possible one-hot encodings on an unseen input. Since the network has been trained to produce high goodness values only for correct labels, the encoding that results in the highest total goodness across all layers is selected as the predicted class. \\\\
We identify two key limitations of FF. Firstly, at inference time, the optimal form of FF demands $C$ separate forward passes, whereas traditional BP-based models require only a single pass. Although \cite{hinton2022forwardforwardalgorithmpreliminaryinvestigations} proposed a more efficient FF variant using a sole forward pass, this version obtained lower accuracy. Secondly, FF has no explicit mechanism to recognize the semantic role of the prepended encoded labels, adding additional complexity to the input representation. To address these issues, we fundamentally redesign the traditional FF framework. Our approach eliminates the need for one-hot encoding and realizes single-pass inference across all classes without increasing the model complexity.

\section{Methodology}
In this research, we redefine the objective in the FF algorithm to a similarity-based objective rather than the goodness score. Instead of the traditional FF formulation's approach of feeding one-hot encoded data through the network twice, we present the model with an \textit{anchor} image, a corresponding \textit{positive} image belonging to the same class as the anchor image, and one or more \textit{negative} images belonging to different classes from the anchor image. Learning is guided by a similarity-based loss function that incentivizes the network to produce embeddings where the anchor is closer to the positive than the negative(s). Subsequently, the loss is calculated based on the embeddings from each layer, and the activations are fed-forward to the next layer to be trained similarly (see \textit{Layer Set-Up} in Fig. \ref{fig:architecture}). Layer normalization is applied such that only the relative magnitudes are passed through. Essentially, layers are trained in a greedy layer-wise fashion without BP. This framework is motivated by the notion that deeper layers can progressively learn more informative representations from previous layers. We present multiple ways of formulating the similarity learning objective and classifying inputs using the learned embeddings.

\subsection{Loss Formulation}
\label{sec:loss-formulation}
\textbf{Triplet Margin Loss.} Given a triplet of input samples $(\mathbf{x},\mathbf{x}^+,\mathbf{x}^-)$, containing an \textit{anchor}, a \textit{positive} image from the same class as the anchor, and a \textit{negative} image from a different class, the triplet margin loss is defined as 
\begin{equation}
    \mathcal{L}_{tri}(\mathbf{f},\mathbf{f}^+,\mathbf{f}^-) =\max(d(\mathbf{f},\mathbf{f}^+)-d(\mathbf{f},\mathbf{f}^-)+\alpha, 0),
    \label{eqn:triplet-loss}
\end{equation}
where $\alpha$ is a margin parameter, $d(\mathbf{a},\mathbf{b})$ is a distance function, and $\mathbf{f}=f(\mathbf{x})$ is a transformation from the input to the embedding space. The Euclidean distance, ${d(\mathbf{a},\mathbf{b})}=||\mathbf{a}-\mathbf{b}||^2$, is used to measure similarity. Notably, the triplet loss function only maximizes the distance between the anchor and one negative class at a time without accounting for other negative classes.

$\boldsymbol{(N+1)}$\textbf{-tuplet Loss.} To overcome the above limitation, we generalize the triplet loss function to an ${(N+1)}$-tuplet formulation \cite{NIPS2016_6b180037}, consisting of an \textit{anchor} image with one \textit{positive} image and $N-1$ \textit{negative} images, each belonging to a different class. Given a tuplet, $(\mathbf{x},\mathbf{x}^+,\{\mathbf{x}^-_i\}_{i=1}^{N-1})$, ${(N+1)}$-tuplet loss is defined as
\begin{multline}
    \mathcal{L}_{tup}(\mathbf{f},\mathbf{f}^+,\{\mathbf{f}_i^-\}_{i=1}^{N-1})\\
    =\log(1+\sum_{i=1}^{N-1}\exp(d(\mathbf{f},\mathbf{f}^+)-d(\mathbf{f},\mathbf{f}^-_i))).
    \label{eqn:tuplet-loss}
\end{multline}
\\
\subsection{FAUST: Vanilla Approach}
\label{sec:vanilla-sec}
Our initial endeavor involves the following framework.

\textbf{Training.}
We begin by utilizing the traditional triplet loss function defined in Section \ref{sec:loss-formulation}---this approach is given the code name \textit{FAUST-vanilla triplet}. Since triplet loss is limited to one negative class at a time, we further adopt the ${(N+1)}$-tuplet formulation described in Section \ref{sec:loss-formulation}---this approach is given the code name \textit{FAUST-vanilla tuplet}. 
${(N+1)}$-tuplet loss, however, introduces added computational complexity from necessitating $(N+1)\times B$ forward passes per training step, where $B$ is the batch size. It is worth noting that, for both formulations above, positive and negative sampling strategies strongly impact the quality of the learned representations. Typical approaches of mining hard examples include the semi-hard criterion \cite{schroff2015facenet}, but these incur higher computation costs. Therefore, for efficiency, we limit our analysis by using only random sampling in \textit{FAUST-vanilla triplet} and \textit{FAUST-vanilla tuplet}. The resulting runtime complexity is ${\mathcal{O}(MC)}$, where $M$ is the size of the training set, and $C$ is the number of classes.

\textbf{Inference.}
We approximate a representation for each class in the embedding space at each layer by randomly sampling a set of images from each class and averaging the computed embeddings to compute a centroid. An input is then classified by finding the class centroid closest to the input's embedding across all layers. The class centroids are computed only once and then cached. The inferred class is defined as
\begin{equation}
\hat{y}(\mathbf{x})=\arg\min_{c\in\mathcal{C}} \sum_{i=1}^L||f_i(\mathbf{x})-\hat{\mathbf{f}}^{c}_{i}||_2,
\end{equation}
where ${\hat{\mathbf{f}}^c_i}$ denotes the approximate class centroid for class ${c}$ and layer $i$. This inference mechanism is considerably more efficient than traditional FF. The optimal form of FF requires $N$ separate forward passes, one for each class hypothesis---our approach only requires a single forward pass per test input.

\begin{table*}[t]
\centering
\renewcommand{\arraystretch}{1.2}
\setlength{\tabcolsep}{10pt}
\begin{tabular}{lcccccc}
\toprule
\textbf{Method} & \multicolumn{2}{c}{MNIST} & \multicolumn{2}{c}{Fashion-MNIST} & \multicolumn{2}{c}{CIFAR-10} \\
\cmidrule(lr){2-3} \cmidrule(lr){4-5} \cmidrule(lr){6-7}
 & 3/500 & 4/800 & 3/500 & 4/800 & 3/500 & 4/800 \\
\midrule
\textbf{Backpropagation}   & 98.64 & 98.64 & 90.11 & 90.40 & 57.30 & 57.63 \\
\midrule
\textbf{Forward-Forward}   & 97.05 & 97.22 & 87.10 & 87.47 & 46.13 & 46.44 \\
\textbf{FFCM\textsuperscript{†}}              & 97.01 & 97.12 & 87.67 & 87.64 & 54.44 & {54.48} \\
\textbf{Collab FF\textsuperscript{†}}         & 97.90 & - & {88.40} & - & 48.40 & - \\
\textbf{FAUST-vanilla triplet}                          & 98.33 & 97.95 & 86.61 & 86.71 & 47.42 & 46.76 \\
\textbf{FAUST-vanilla tuplet}                          & \textbf{98.68} & \textbf{98.69} & 87.69 & 87.42 & 54.05 & 53.79  \\
\textbf{FAUST-representative tuplet}                  & {98.43} & {98.39} & \textbf{89.67} & \textbf{89.47} & \textbf{55.92} & \textbf{56.22} \\
\bottomrule
\end{tabular}
\caption{\textbf{Accuracy across datasets and algorithms.} $M/N$ denotes training the algorithm on $M$ layers of $N$ neurons. Algorithms marked with \textsuperscript{†} have no official source code available, and the best results reported in their original papers are taken \cite{10491157, Lorberbom_Gat_Adi_Schwing_Hazan_2024}.}
\label{tab:method_comparison}
\end{table*}

\subsection{FAUST: Representative Approach}
\label{sec:representative-sec}
The vanilla variations described previously have two crucial limitations: \textit{FAUST-vanilla triplet} accounts for only one negative class at a time, and \textit{FAUST-vanilla tuplet} considerably increases computational complexity. In this approach, we reduce the computational overhead from passing ${N-1}$ negative samples by simplifying the underlying optimization problem. The following algorithm is given the code name \textit{FAUST-representative tuplet}; it presents a more efficient means to exploit the ${(N+1)}$-tuplet loss function. \\\\
We extract one image from each class to globally represent the class across training and inference (\textit{Representatives} in Fig. \ref{fig:architecture}). This set of representatives is denoted $\mathcal{R}$, where $|\mathcal{R}|=C$ is the number of classes. Every image within a batch of size $B$ is taken as an anchor. A ${(C+1)}$-tuplet is formed comprising this anchor, the corresponding positive image from $\mathcal{R}$, and the remaining negative images $\mathcal{R}\backslash\{\mathbf{x}^+\}$. We apply the ${(N+1)}$-tuplet loss using these tuplets. At the beginning of each training batch, the embeddings for all representative images are computed and cached as fixed reference points for the entire batch, eliminating the need for recomputations for each anchor (see \textit{Embedding Cache} and \textit{Training Phase} in Fig. \ref{fig:architecture}). This setup requires ${B+C}$ forward passes per training batch, which is strictly less than the $3B$ passes required for triplet loss with random sampling while maintaining a linear runtime complexity. \\\\
The use of fixed representatives simplifies the optimization problem from minimizing the loss over all possible tuplet combinations---with varied positives and negatives for a given anchor---to a simpler subproblem involving only fixed positives and negatives. Therefore, the representative-based formulation is a constrained variant of the full tuplet-based objective. \\\\
We modify the inference method correspondingly for this redefined problem. The inference phase of \textit{FAUST-representative tuplet} is similar to that of \textit{FAUST-vanilla triplet} and \textit{FAUST-vanilla tuplet}, barring one vital difference---instead of taking an average, each representative image is propagated through the trained network, and their embeddings are cached directly. As before, the input image is then classified as the class of the representative image with the minimum Euclidean distance.\\\\
Overall, this algorithm is defined in Algorithm \ref{algorithm-1}.

\subsection{Embedding Layer}
We observe that mapping each layer's activations to a lower-dimensional embedding space through a trainable linear projection significantly improves the quality of the representations compared to an identity mapping, a result consistent with \cite{2002.05709}. The loss function is optimized on these embeddings, but the activations prior to the embedding layer are passed onto subsequent layers to maximize the information flow through the network. The activations are represented as $\mathbf{g}_{i+1}=\phi(W_1^{(i+1)}\mathbf{g}_i)$, and the embeddings are represented as $\mathbf{f}_{i+1}=W_2^{(i+1)}\mathbf{g}_{i+1}$, where $W_1$ and $W_2$ denote the weight matrices for the main fully connected layer and the embedding layer, respectively.
\begin{algorithm}[h]
\caption{Training in \textit{FAUST-representative tuplet}\\
for one batch}
\begin{algorithmic}[1]
\setstretch{1.2}
\State \textbf{Input:} $\mathbf{x},\; \mathbf{r}$ \Comment{Input and Representatives}
\For{$i \leftarrow 1, \text{num\_layers}$}
    \State $\mathbf{g}^{\mathbf{r}}_{i}=\phi(W_1^{(i)}\mathbf{r})$,
    \State $\mathbf{f}^{\mathbf{r}}_i \gets =W_2^{(i)}\mathbf{g}^{\mathbf{r}}_{i}$ 
    \Comment{Embedding Cache}
    \State $\mathbf{g}_{i}=\phi(W_1^{(i)}\mathbf{x})$,
    \State $\mathbf{f}_i \gets =W_2^{(i)}\mathbf{g}_i$ 
    \Comment{Input Embedding}
    \State $\mathcal{L}_i \gets \text{Tuplet}(\mathbf{f}_i, \mathbf{f}^{\mathbf{r}}_{i})$
    \State $W_1^{(i)} \gets W_1^{(i)} - \eta \frac{\partial \mathcal{L}_i}{\partial W_1^{(i)}}$
    \State $W_2^{(i)} \gets W_2^{(i)} - \eta \frac{\partial \mathcal{L}_i}{\partial W_2^{(i)}}$
    \State Pass $\mathbf{x} = \mathbf{f}_i,\mathbf{r} = \mathbf{f}^\mathbf{r}_i$ to next layer with detached gradient
\EndFor
\end{algorithmic}
\label{algorithm-1}
\end{algorithm}
\vspace{-0.001em}
\section{Implementation}
Our algorithm is tested on an MLP containing a stack of hidden layers, each comprising a fully connected layer followed by a ReLU activation, layer normalization, and an embedding layer. We configure the network to have either three layers and 500 neurons per layer or four layers and 800 neurons per layer. We fine-tune hyperparameters such as the optimizer, batch size, and learning rate for stable convergence. The three variations of our algorithm from the previous section are implemented: %
\begin{inparaenum}[\itshape i.\upshape]
  \item \textit{FAUST-vanilla triplet},
  \item \textit{FAUST-vanilla tuplet}, and
  \item \textit{FAUST-representative tuplet}.
\end{inparaenum}

\section{Results}
\vspace{-0.1em}
\captionsetup[table]{position=bottom, justification=raggedright, singlelinecheck=false}

\begin{figure*}[h]
    \centering
    \begin{subfigure}[t]{0.24\textwidth}
        \includegraphics[width=\linewidth]{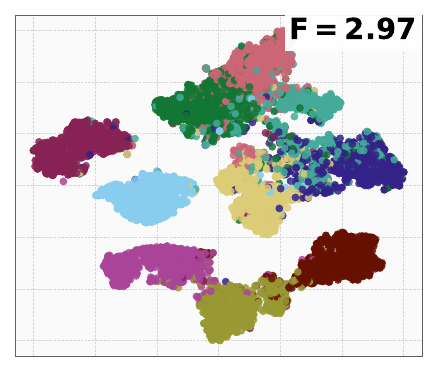}
        \caption{Layer 1}
    \end{subfigure}
    \hfill
    \begin{subfigure}[t]{0.24\textwidth}
        \includegraphics[width=\linewidth]{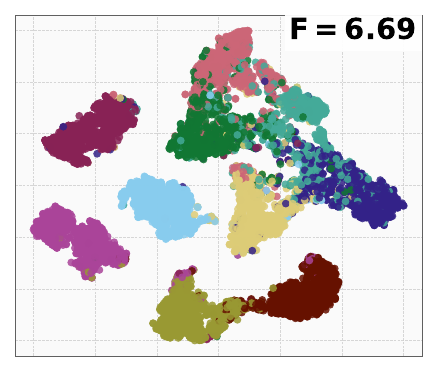}
        \caption{Layer 2}
    \end{subfigure}
    \hfill
    \begin{subfigure}[t]{0.24\textwidth}
        \includegraphics[width=\linewidth]{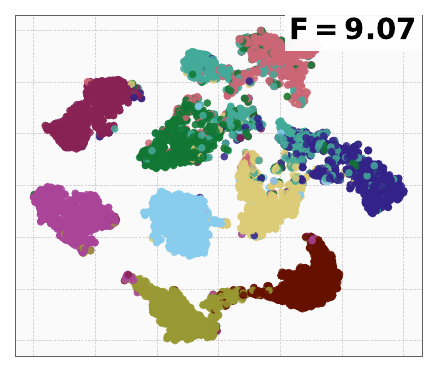}
        \caption{Layer 3}
    \end{subfigure}
    \hfill
    \begin{subfigure}[t]{0.24\textwidth}
        \includegraphics[width=\linewidth]{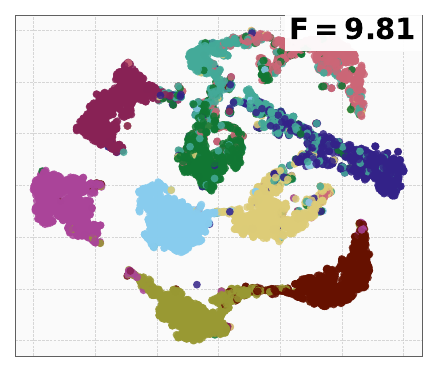}
        \caption{Layer 4}
    \end{subfigure}
    \caption{$\boldsymbol{t}$\textbf{-SNE visualizations} of the representations learned by \textit{FAUST-representative tuplet} on Fashion-MNIST, using an MLP with four layers of 500 neurons and an embedding size of 256. The value in the upper-right corner, $F$, is the Fisher discriminant score. Color-to-class mapping is (\textcolor{clsTshirt}{T-shirt/top},
\textcolor{clsTrouser}{Trouser},
\textcolor{clsPullover}{Pullover},
\textcolor{clsDress}{Dress},
\textcolor{clsCoat}{Coat},
\textcolor{clsSandal}{Sandal},
\textcolor{clsShirt}{Shirt},
\textcolor{clsSneaker}{Sneaker},
\textcolor{clsBag}{Bag},
\textcolor{clsAnkleboot}{Ankle-boot}).}
    \label{fig:tsne-plot}
\end{figure*} 

\begin{figure*}[htbp]
    \centering
    \begin{subfigure}[t]{0.30\textwidth}
        \includegraphics[width=\linewidth]{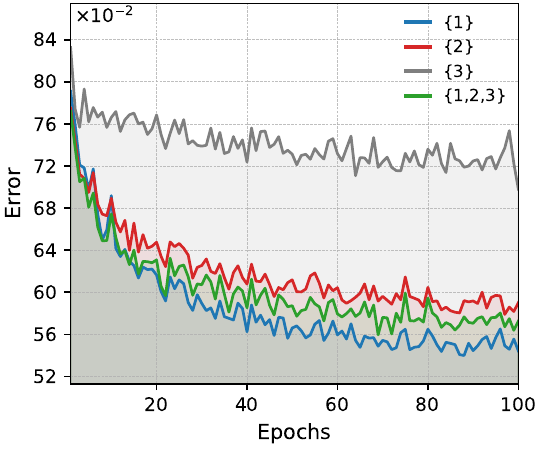}
        \caption{FF}
    \end{subfigure}
    \hfill
    \begin{subfigure}[t]{0.30\textwidth}
        \includegraphics[width=\linewidth]{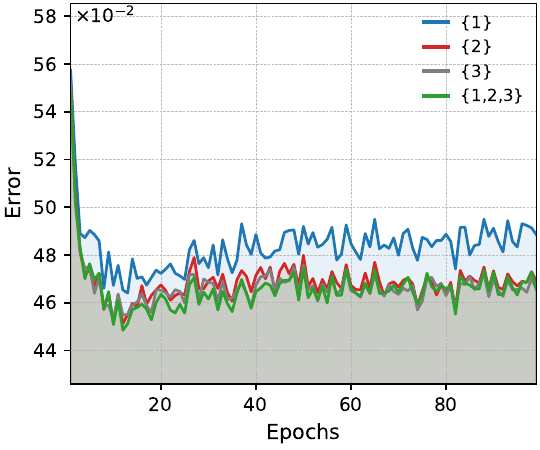}
        \caption{FAUST}
    \end{subfigure}
    \hfill
    \begin{subfigure}[t]{0.30\textwidth}
        \includegraphics[width=\linewidth]{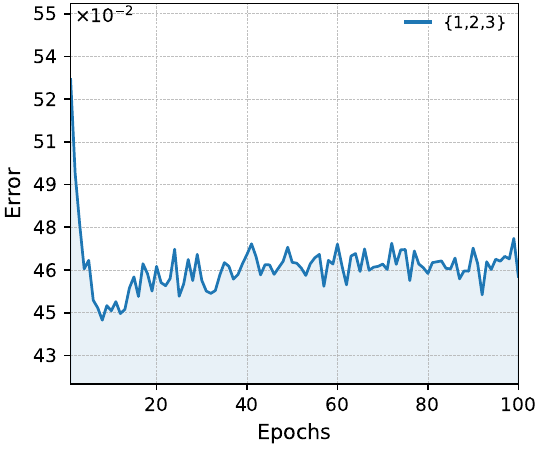}
        \caption{BP}
    \end{subfigure}
    \caption{\textbf{Convergence curves} of FF, \textit{FAUST}, and BP on CIFAR-10, using an MLP with three layers of 500 neurons and an embedding size of 256. All models were trained with the Adam optimizer (learning rate = 0.001, batch size = 256). In the legend, \{1\} denotes predictions from layer 1 alone, and \{1,2,3\} represents the combined predictions from all three layers. Note that BP does not provide layer-wise predictions.}
    \label{fig:convergence-plot}
\end{figure*}

\subsection{Classification Accuracy}
We train and evaluate our network on three datasets: MNIST, Fashion-MNIST, and CIFAR-10. Table \ref{tab:method_comparison} presents our experimental results compared against the benchmarks set by BP, FF, Collab FF \cite{Lorberbom_Gat_Adi_Schwing_Hazan_2024}, and FFCM \cite{10491157}.  \\\\
Across all datasets, \textit{FAUST-vanilla tuplet} achieves a better accuracy of up to 7.03\% higher than \textit{FAUST-vanilla triplet}, confirming the effectiveness of extending the range of comparison through the ${(N+1)}$-tuplet loss function. On MNIST, \textit{FAUST-vanilla tuplet} and \textit{FAUST-representative tuplet} perform similarly. There are, however, notable differences on Fashion-MNIST and CIFAR-10, with \textit{FAUST-representative tuplet} outperforming \textit{FAUST-vanilla tuplet}. This improvement appears to confirm our hypothesis that fixed representatives create a simpler optimization problem that results in better classification performance.\\\\

We compare \textit{FAUST} against previously published FF and BP algorithms. \textit{FAUST-representative tuplet} outperforms FF, FFCM, and Collab FF across all three datasets. On MNIST and Fashion-MNIST, \textit{FAUST-representative tuplet} achieves similar accuracies to BP, with the difference being no greater than 0.93\%. On CIFAR-10, while BP continues to surpass all BP-free algorithms, \textit{FAUST-representative tuplet} reduces the gap to these BP benchmarks.

\subsection{Behavior of Deeper Layers}
We visualize the embeddings produced by each layer of \textit{FAUST-representative tuplet} on Fashion-MNIST with the $t$-distributed stochastic neighbor embedding ($t$-SNE) method in Figure \ref{fig:tsne-plot}. The 10 clusters demonstrate that the embeddings are discriminative across all labels of the dataset. While each layer is trained independently, the discriminative power increases in deeper layers. We quantify this increase with the Fisher discriminant score, which increases from F = 2.97 to F = 9.81 across four layers. This result demonstrates that the representations learned by shallower layers can provide useful information for learning more abstract embeddings in deeper layers. To further analyze the behavior of \textit{FAUST} compared to FF and BP, we plot the convergence curves in Figure \ref{fig:convergence-plot}. On CIFAR-10, FF demonstrates poor scalability, as deeper layers fail to reduce classification error. In contrast, \textit{FAUST} demonstrates a convergence rate comparable to BP.

\section{Conclusions}
In this research, we present a novel BP-free algorithm that redefines FF within a similarity learning paradigm. Our algorithm, \textit{FAUST}, achieves promising results and requires only one forward pass for each inference. It is found that \textit{FAUST} outperforms existing FF algorithms on the evaluated benchmarks by a considerable margin. While BP continues to achieve better performance on complex datasets, \textit{FAUST} narrows the gap to these BP benchmarks. Notably, on CIFAR-10, \textit{FAUST} achieves 56.22\% accuracy, comparable to the BP benchmark of 57.63\% accuracy. The natural progression would be to apply \textit{FAUST} to more challenging tasks and larger networks including CNNs. While selecting one representative image from each class produces compelling results on the evaluated datasets, other methods of formulating the similarity objective may also be worth exploring.

\section*{Author Contribution}

James Gong, Raymond Luo, Emma Wang, and Leon Ge contributed equally to this work.

\bibliographystyle{IEEEtran}
\bibliography{mybib}
\end{document}